\documentclass{article} % For LaTeX2e
\usepackage{nips14submit_e,times}
\usepackage{hyperref}
\usepackage{url}
\usepackage[numbers,sort]{natbib}
\usepackage{graphicx}
\usepackage{amsmath,amssymb}
\usepackage{array}
\usepackage{multirow}
\usepackage{color,colortbl}
\usepackage[compact]{titlesec}
\usepackage{array}
\usepackage{enumitem}

\nipsfinalcopy % Uncomment for camera-ready version

\setenumerate{noitemsep,topsep=0pt,parsep=0pt,partopsep=0pt,labelindent=0pt,leftmargin=1.2em}
\titlespacing{\section}{0pt}{0pt}{0pt}
\titlespacing{\subsection}{0pt}{0pt}{0pt}
\newcommand{\figref}[1]{Fig.~\ref{#1}}
\newcommand{\tblref}[1]{Table~\ref{#1}}
\newcommand{\sref}[1]{Sect.~\ref{#1}}
\def\eg{\emph{e.g.}}
\def\ie{\emph{i.e.}}

\def\etal{\emph{et al.}}

\definecolor{Gray}{gray}{0.95}

\newcommand{\dict}{\mathcal{W}}
\newcommand{\chars}{\mathcal{C}}

\renewcommand{\paragraph}[1]{\par\noindent{\bf #1}}

\begin{document}
\title{Synthetic Data and Artificial Neural Networks for\\ Natural Scene Text Recognition}
\author{
Max Jaderberg \\
\And
Karen Simonyan \\
\And
Andrea Vedaldi \\
\And
Andrew Zisserman
}
\maketitle
 \vspace{-3.5em}
 \begin{center}
 Visual Geometry Group, University of Oxford\\
 \texttt{\{max,karen,vedaldi,az\}@robots.ox.ac.uk}  
 \end{center}
 \vspace{1em}
\begin{abstract}
In this work we present a framework for the recognition of natural scene text. Our framework does not require any human-labelled data, and performs word recognition on the whole image holistically, departing from the character based recognition systems of the past. The deep neural network models at the centre of this framework are trained solely on data produced by a synthetic text generation engine -- synthetic data that is highly realistic and sufficient to replace real data, giving us infinite amounts of training data. This excess of data exposes new possibilities for word recognition models, and here we consider three models, each one ``reading'' words in a different way: via 90k-way dictionary encoding, character sequence encoding, and bag-of-N-grams encoding. In the scenarios of language based and completely unconstrained text recognition we greatly improve upon state-of-the-art performance on standard datasets, using our fast, simple machinery and requiring zero data-acquisition costs.

%In this work we present a framework for the recognition of natural scene text. We use purely data-driven, deep learning models to perform word recognition on the whole image at the same time, departing from the character based recognition systems of the past. These models are trained solely on data produced by a synthetic text generation engine -- synthetic data that is highly realistic and sufficient to replace real data, giving us infinite amounts of training data. This excess of data exposes new possibilities for word recognition models, and here we introduce three novel models, each one ``reading'' words in a complementary way: via large-scale dictionary encoding, character sequence encoding, and bag-of-N-gram encoding. In the scenarios of language/lexicon based and completely unconstrained text recognition we demonstrate state-of-the-art performance on standard datasets, using our fast, simple machinery and requiring zero data-acquisition costs.
\end{abstract}

% --------------------------------------------------------------
% --------------------------------------------------------------
\section{Introduction}
% --------------------------------------------------------------
% --------------------------------------------------------------
Text recognition in natural images, \emph{scene text recognition}, is a challenging but wildly useful task. Text is one of the basic tools for preserving and communicating information, and a large part of the modern world is designed to be interpreted through the use of labels and other textual cues. This makes scene text recognition imperative for many areas in information retrieval, in addition to being crucial for human-machine interaction.

While the recognition of text within scanned documents is well studied and there are many document OCR systems that perform very well, these methods do not translate to the highly variable domain of scene text recognition. When applied to natural scene images, traditional OCR techniques fail as they are tuned to the largely black-and-white, line-based environment of printed documents, while text occurring in natural scene images suffers from inconsistent lighting conditions, variable fonts, orientations, background noise, and imaging distortions.

To effectively recognise scene text, there are generally two stages: word detection and word recognition. The detection stage generates a large set of word bounding box candidates, and is tuned for speed and high recall. Previous work uses sliding window methods~\cite{Wang11} or region grouping methods~\cite{Chen11,Epshtein10,Neumann12} very successfully for this. Subsequently, these candidate detections are recognised, and this recognition process allows for filtering of false positive word detections. Recognition is therefore a far more challenging problem and it is the focus of this paper.

%We make two major contributions in this work. Firstly, we introduce three models that give state-of-the-art scene text recognition performance through the use of large, supervised convolutional neural networks (CNNs)~\cite{Lecun98}. These models have diverse representations of words, suitable for different purposes: a novel large scale dictionary encoding model~(\sref{sec:dictionary}), a character sequence encoding model~(\sref{sec:characters}), and a novel bag-of-N-grams encoding model~(\sref{sec:ngrams}). Our recognition methods work by performing inference on the word image \emph{holistically}, mimicking more closely how humans read~\cite{Coltheart01}, rather than by \emph{sequential} character classification as is traditionally done. Secondly, we meet the huge training data requirements of these models with a synthetic data generation method. Our extensive evaluation shows that we can generate sufficiently realistic data without a human annotator, requiring only an unlabelled natural image dataset and a font collection, yet still obtain state-of-the-art results on real-world data. This allows our framework to be seamlessly extended to larger vocabularies and other languages without any human-labelling cost.

While most approaches recognize individual characters by pooling evidence locally, Goodfellow~\etal~\cite{Goodfellow13} do so from the image of the whole character string using a convolutional neural network (CNN)~\cite{Lecun98}. They apply this to street numbers and synthetic CAPTCHA recognition obtaining excellent results. Inspired by this approach, we move further in the direction of holistic word classification for scene text, and make two important contributions. Firstly, we propose a state-of-the-art CNN text recogniser that also pools evidence from images of entire words. Crucially, however, we regress all the characters simultaneously, formulating this as a classification problem in a large lexicon of 90k possible words~(\sref{sec:dictionary}). In order to do so, we show how CNNs can be efficiently trained to recognise a very large number of words using \emph{incremental training}. While our lexicon is restricted, it is so large that this hardly constitutes a practical limitation. Secondly, we show that this state-of-the-art recogniser can be trained \emph{purely from synthetic data}. This result is highly non-trivial as, differently from CAPTCHA, the classifier is then applied to real images. While synthetic data was used previously for OCR, it is remarkable that this can be done for scene text, which is significantly less constrained. This allows our framework to be seamlessly extended to larger vocabularies and other languages without any human-labelling cost.
In addition to these two key contributions, we study two alternative models -- a character sequence encoding model with a modified formulation to that of~\cite{Goodfellow13}~(\sref{sec:characters}), and a novel bag-of-N-grams encoding model which predicts the unordered set of N-grams contained in the word image~(\sref{sec:ngrams}).

A discussion of related work follows immediately and our data generation system described after in~\sref{sec:synth}. Our deep learning word recognition architectures are presented in~\sref{sec:architectures}, evaluated in~\sref{sec:eval}, and conclusions are drawn in~\sref{sec:conclusions}.

% --------------------------------------------------------------
% --------------------------------------------------------------
\paragraph{Related work.}\label{sec:related}
% --------------------------------------------------------------
% --------------------------------------------------------------
%We concentrate here on text recognition methods, given a bounding box
%in the image, rather than the text detections stage of scene text
%recognition.  
%
Traditional text recognition methods are based on sequential character
classification by either sliding
windows~\cite{Wang11,Wang12,Jaderberg14a} or connected
components~\cite{Neumann10a,Neumann12}, after which a word
prediction is made by grouping character classifier predictions in a
left-to-right manner. The sliding window classifiers include random
ferns~\cite{Ozuysal07} in Wang~\etal~\cite{Wang11}, and CNNs in~\cite{Wang12,Jaderberg14a}. Both~\cite{Wang11} and~\cite{Wang12} use a
small fixed lexicon as a language model to constrain word recognition.

More recent works such as~\cite{Bissacco13,Neumann13,Alsharif13} make
use of over-segmentation methods, guided by a supervised classifier, to
generate candidate proposals which are subsequently classified as
characters or false positives. For example, PhotoOCR~\cite{Bissacco13} uses binarization and a sliding window classifier to generate candidate character regions, with words recognised through a beam search driven by classifier scores followed by a re-ranking using a dictionary of 100k words.~\cite{Jaderberg14a} uses the convolutional nature of CNNs to generate response maps for characters and bigrams which are integrated to score lexicon words.

%Methods of generating candidate
%character regions include: oriented stroke detection~\cite{Neumann13},
%binarization and a sliding window classifier as in the PhotoOCR system
%of~\cite{Bissacco13} (which uses a dictionary of 100k words for
%re-ranking candidate words), and a CNN character classifier in
%Alsharif~\etal~\cite{Alsharif13} (which subsequently uses a 50k dictionary for
%accurate recognition). Words are recognised through a sequential beam search optimization over character candidates.

In contrast to these approaches based on character
classification, the work
by~\cite{Goel13,Rodriguez13,Novikova12,Mishra12} instead uses the
notion of holistic word recognition.~\cite{Mishra12,Novikova12} still
rely on explicit character classifiers, but construct a graph to infer
the word, pooling together the full word
evidence. Rodriguez~\etal~\cite{Rodriguez13} use aggregated Fisher
Vectors~\cite{Perronnin10} and a Structured SVM
framework to create a joint word-image and text
embedding.~\cite{Goel13} use whole word-image features to recognize
words by comparing to simple black-and-white font-renderings of
lexicon words.

Goodfellow~\etal~\cite{Goodfellow13} had great success using a CNN with multiple position-sensitive character classifier outputs (closely related to the character sequence model in~\sref{sec:characters}) to perform street number recognition. This model was extended to CAPTCHA sequences (up to 8 characters long) where they demonstrated impressive performance using synthetic training data for a synthetic problem (where the generative model is known), but we show that synthetic training data can be used for a real-world data problem (where the generative model is unknown).

%While not performing full text recognition, the system presented by
%Goodfellow~\etal~\cite{Goodfellow13} recognises street numbers from
%loosely cropped images. They use a large CNN which outputs the
%sequence of numbers in the image almost directly. This is also applied
%to full text CAPTCHAs with great success, however, although randomly
%generated, CAPTCHAs are purely synthetic images. Their model is
%closely related to the character sequence encoding model
%in~\sref{sec:characters}, with similarities discussed in that
%section.

% In this work, we move further in the direction of holistic word classification for scene text. Our first model (\sref{sec:dictionary}) performs word classification in a standard CNN object recognition manner across 90k classes, made possible by \emph{incremental training}. The second model (\sref{sec:characters}), similar to~\cite{Goodfellow13}, predicts directly the sequence of characters contained within the image. Our third method (\sref{sec:ngrams}) performs N-gram classification, creating a bag-of-N-grams embedding that can be easily used to infer the full word. All three models take the whole word image as input to perform recognition.

% --------------------------------------------------------------
% --------------------------------------------------------------
\section{Synthetic Data Engine}\label{sec:synth}
% --------------------------------------------------------------
% --------------------------------------------------------------
\begin{figure}[t]
\centering
\begin{tabular}{>{\centering}m{10pt}m{0.8\textwidth}}
(a) & \includegraphics[width=0.8\textwidth]{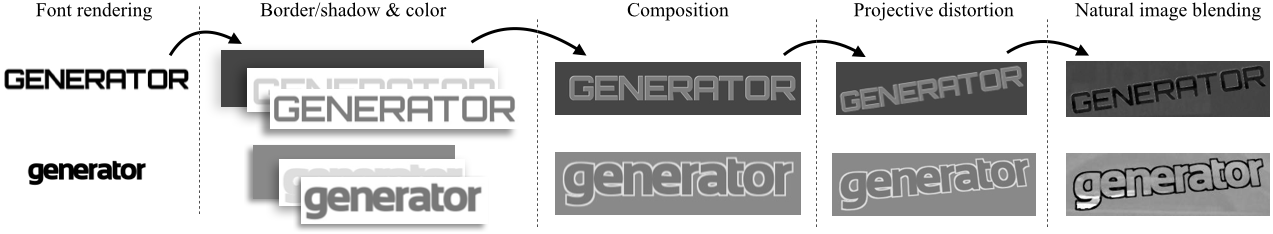}\\
\hline
(b) & \includegraphics[width=0.8\textwidth,trim=0 0 0 -10pt]{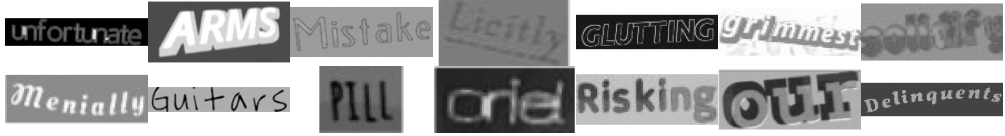}\\
\end{tabular}
\vspace*{-3mm}
\caption{(a) The text generation process after font rendering, creating and coloring the image-layers, applying projective distortions, and after image blending. (b) Some randomly sampled data created by the synthetic text engine.}
\label{fig:synthdata}
\end{figure}

This section describes our scene text rendering algorithm. As our CNN models take whole word images as input instead of individual character images, it is essential to have access to a training dataset of cropped word images that covers the whole language or at least a target lexicon. While there are some publicly available datasets from ICDAR~\cite{ICDAR03, ICDAR2005, ICDAR11, ICDAR2013}, the Street View Text (SVT) dataset~\cite{Wang11} and others, the number of full word image samples is only in the thousands, and the vocabulary is very limited. These limitations have been mitigated before by mining for data or having access to large proprietary datasets~\cite{Jaderberg14a,Bissacco13}, but neither of these approaches are wholly accessible or scalable.

%The lack of full word image samples has caused previous work to rely on character recognition instead (as character data is plentiful) or the deficit has been mitigated by mining for data or having access to large proprietary datasets~\cite{Bissacco13,Goodfellow13}, but neither of these approaches are wholly accessible or scalable.

Here we follow the success of some synthetic character datasets~\cite{Campos09,Wang12} and create a synthetic word data generator, capable of emulating the distribution of scene text images. This is a reasonable goal, considering that much of the text found in natural scenes is computer-generated and only the physical rendering process (\eg~printing, painting) and the imaging process (\eg~camera, viewpoint, illumination, clutter) are not controlled by a computer algorithm. 
% The key to our synthetic text generator is that all the components of the synthetic image are blended to resemble natural images, keeping the data distribution in line with that of the real world data.

\figref{fig:synthdata} illustrates the generative process and some resulting synthetic data samples. These samples are composed of three separate image-layers -- a background image-layer, foreground image-layer, and optional border/shadow image-layer -- which are in the form of an image with an alpha channel. The synthetic data generation process is as follows:
\begin{enumerate}
  \item {\emph{Font rendering} -- a font is randomly selected from a catalogue of over 1400 fonts downloaded from Google Fonts. The kerning, weight, underline, and other properties are varied randomly from arbitrarily defined distributions. The word is rendered on to the foreground image-layer's alpha channel with either a horizontal bottom text line or following a random curve.}
  \item \emph{Border/shadow rendering} -- an inset border, outset border or shadow with a random width may be rendered from the foreground.

\item \emph{Base coloring} -- each of the three image-layers are 
filled with a different uniform color sampled
from clusters over natural images. The clusters are formed by
k-means clustering the three color components of each image of the
training datasets of~\cite{ICDAR03} into three clusters.

  \item \emph{Projective distortion} -- the foreground and border/shadow image-layers are distorted with a random, full-projective transformation, simulating the 3D world.
  \item \emph{Natural data blending} -- each of the image-layers are blended with a randomly-sampled crop of an image from the training datasets of ICDAR 2003 and SVT. The amount of blend and alpha blend mode (\eg~normal, add, multiply, burn, max, \emph{etc.}) is dictated by a random process, and this creates an eclectic range of textures and compositions. The three image-layers are also blended together in a random manner, to give a single output image.
  \item \emph{Noise} -- Gaussian noise, blur, and JPEG compression artefacts are introduced to the image.
\end{enumerate}

The word samples are generated with a fixed height of 32~pixels, but with a variable width. Since the input to our CNNs is a fixed-size image, the generated word images are rescaled so that the width equals 100 pixels.
Although this does not preserve the aspect ratio, the horizontal frequency distortion of image features most likely provides the word-length cues. We also experimented with different padding regimes to preserve the aspect ratio, but found that the results are not quite as good as with resizing. 

The synthetic data is used in place of real-world data, and the labels are generated from a corpus or dictionary as desired. By creating training datasets much larger than what has been used before, we are able to use data-hungry deep learning algorithms to train richer, whole-word-based models.

% --------------------------------------------------------------
% --------------------------------------------------------------
\section{Models}\label{sec:architectures}

In this section we describe three models for visual recognition of scene text words.
All use the same framework of generating synthetic text data (\sref{sec:synth}) to train deep convolutional networks on whole-word image samples, but with different objectives, which correspond to different methods of reading. \sref{sec:dictionary} describes a model performing pure word classification to a large dictionary, explicitly modelling the entire known language. \sref{sec:characters} describes a model that encodes the character at each position in the word, making no language assumptions to naively predict the sequence of characters in an image. \sref{sec:ngrams} describes a model that encodes a word as a bag-of-N-grams, giving a compositional model of words as not only a collection of characters, but of 2-grams, 3-grams, and more generally, N-grams. 

\subsection{Encoding Words}\label{sec:dictionary}
This section describes our first model for word recognition, where words $w$ are constrained to be selected in a pre-defined dictionary $\dict$. We formulate this as multi-class classification problem, with one class per word. While the dictionary $\dict$ of a natural language may seem too large for this approach to be feasible, in practice an advanced English vocabulary, including different word forms, contains only around 90k words, which is large but manageable.

\begin{figure}[t]
\centering
\vspace{-1em}
(a)
\includegraphics[width=0.75\textwidth]{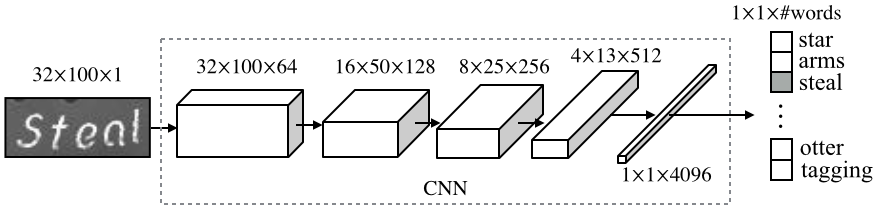}
\begin{tabular}{cccc}
\includegraphics[width=0.55\textwidth]{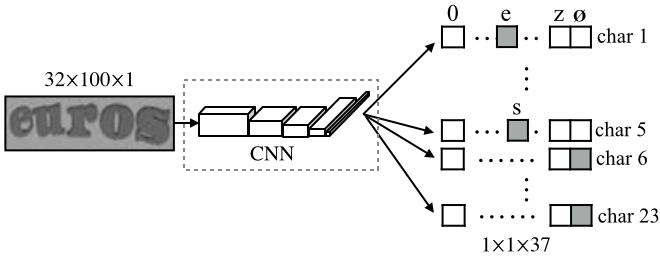}&
\includegraphics[width=0.4\textwidth]{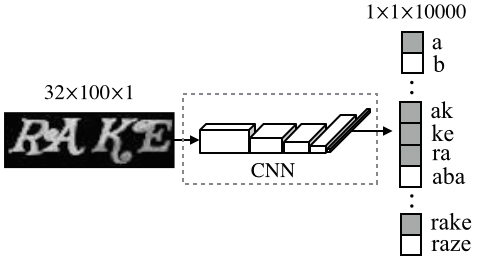}
\vspace{-2em}
\\
(b)&(c)
\end{tabular}
\vspace{-1em}
% \vspace*{-1mm}
\caption{A schematics of the CNNs used showing the dimensions of the featuremaps at each stage for (a) dictionary encoding, (b) character sequence encoding, and (c) bag-of-N-gram encoding. The same five-layer, base CNN architecture is used for all three models.}
\label{fig:dictnet}
\end{figure}

In detail, we propose to use a CNN classifier where each word $w\in\dict$ in the lexicon corresponds to an output neuron. We use a CNN with four convolutional layers and two fully connected layers. Rectified linear units are used throughout after each weight layer except for the last one. In forward order, the convolutional layers have 64, 128, 256, and 512 square filters with an edge size of 5, 5, 3, and 3. Convolutions are performed with stride 1 and there is input feature map padding to preserve spatial dimensionality. $2 \times 2$ max-pooling follows the first, second and third convolutional layers. The fully connected layer has 4096 units, and feeds data to the final fully connected layer which performs classification, so has the same number of units as the size of the dictionary we wish to recognize. The predicted word recognition result $w^*$ out of the set of all dictionary words $\dict$ in a language $\mathcal{L}$ for a given input image $x$ is given by $w^* = \arg\max_{w \in \dict} P(w|x,\mathcal{L})$. Since $P(w|x,\mathcal{L}) = \frac{P(w|x)P(w|\mathcal{L})P(x)}{P(x|\mathcal{L})P(w)}$ and with the assumptions that $x$ is independent of $\mathcal{L}$ and that prior to any knowledge of our language all words are equally probable, our scoring function reduces to $w^* = \arg\max_{w \in \dict} P(w|x)P(w|\mathcal{L})$. The per-word output probability $P(w|x)$ is modelled by the softmax scaling of the final fully connected layer, and the language based word prior $P(w|\mathcal{L})$ can be modelled by a lexicon or frequency counts. A schematic of the network is shown in~\figref{fig:dictnet}~(a).

\paragraph{Training.}
We train the network by back-propagating the standard multinomial logistic regression loss with dropout~\cite{Hinton12}, which improves generalization. Optimization uses stochastic gradient descent (SGD), dynamically lowering the learning rate as training progresses. With uniform sampling of classes in training data, we found the SGD batch size must be at least a fifth of the total number of classes in order for the network to train. 

For very large numbers of classes (\ie~over 5k classes), the SGD batch size required to train effectively becomes large, slowing down training a lot. Therefore, for large dictionaries, we perform \emph{incremental training} to avoid requiring a prohibitively large batch size. This involves initially training the network with 5k classes until partial convergence, after which an extra 5k classes are added. The original weights are copied for the original 5k classes, with the new classification layer weights being randomly initialized. The network is then allowed to continue training, with the extra randomly initialized weights and classes causing a spike in training error, which is quickly trained away. This process of allowing partial convergence on a subset of the classes, before adding in more classes, is repeated until the full number of desired classes is reached. In practice for this network, the CNN trained well with initial increments of 5k classes, and after 20k classes is reached the number of classes added at each increment is increased to 10k.

\subsection{Encoding Sequences of Characters}\label{sec:characters}
This section describes a different model for word recognition. Rather than having a single large dictionary classifier as in~\sref{sec:dictionary}, this model uses a single CNN with multiple independent classifiers, each one predicting the character at each position in the word. 
This character sequence encoding model is a complete departure from the dictionary-constrained model, as this allows entirely unconstrained recognition of words.

A word $w$ of length $N$ is modelled as a sequence of characters such that $w = (c_1, c_2, \dots, c_N)$ where each $c_i \in \chars = \{1,2,\dots,36\}$ represents a character at position $i$ in the word, from the set of 10 digits and 26 letters. Each $c_i$ can be predicted with a single classifier, one for each character in the word. However, since words have variable length $N$ which is unknown at test time, we fix the number of characters to 23, the maximum length of a word in the training set, and introduce a null character class. Therefore a word is represented by a string $w = (\chars \cup \{\phi\})^{23}$. Then for a given image $x$, each character is predicted as $c_i^* = \arg\max_{c_i \in \chars \cup \{\phi\}} P(c_i|\Phi(x))$. $P(c_i|\Phi(x))$ is given by the $i$-th classifier acting on a single set of shared CNN features $\Phi(x)$. 

The base CNN has the same structure as the first five layers of~\sref{sec:dictionary}: four convolutional layers followed by a fully connected layer, giving $\Phi(x)$. The output of the fully connected layer is then fed to 23 separate fully connected layers with 37 neurons each, one for each character class. These fully connected layers are independently softmax normalized and can be interpreted as the probabilities $P(c_i|\Phi(x))$ of the width-resized input image $x$. \figref{fig:dictnet}~(b) illustrates this model.
% \paragraph{Training.}
The model is trained as in~\sref{sec:dictionary} on purely synthetic data by SGD with dropout regularisation, back-propagating gradients from each 23 softmax classifier to the base net.

\paragraph{Discussion.}
This sequential character encoding model is similar to the model used by Goodfellow~\etal~in~\cite{Goodfellow13}. Although the model of~\cite{Goodfellow13} is not applied to scene text (only street numbers and CAPTCHA puzzles), it uses a separate character classifier for each letter in the word, able to recognise numbers up to 5 digits long and CAPTCHAs up to 8 characters long. However, rather than incorporating a no-character class in each character positions's classifier, a further length classifier is trained to output the predicted length of the word. This requires a final post-processing stage to find the optimal word prediction given the character classifier outputs and the length classifier output. We achieve a similar effect but without requiring any post processing -- the word can be read directly from the CNN output, stripping the no-character class predictions.

\subsection{Encoding Bags of N-grams}\label{sec:ngrams}
This section describes our last word recognition model, which exploits
compositionality to represent words. In contrast to the sequential
character encoding of~\sref{sec:characters}, words can be seen as a
composition of an unordered set of character N-grams, a
\emph{bag-of-N-grams}. In the following, if $s\in \chars^N$ and
$w\in\chars^M$ are two strings, the symbol $s \subset w$ indicates
that $s$ is a substring
% \footnote{I.e. $\exists j\in\{1,\dots,N-M+1\} :
% s_1 = w_j \wedge \dots \wedge s_M = w_{j+M-1}.$} 
of $w$. An $N$-gram
of word $w$ is a substring $s \subset w$ of length $|w|=N$. We will
denote with $G_N(w) = \{ s : s\subset w \,\wedge\,|s|\leq N\}$ the set
of all grams of word $w$ of length up to $N$ and with $G_N
=\cup_{w\in\dict} G_N(w)$ the set of all such grams in the
language. For example,
$G_3(\text{spires})=\{\text{s},\text{p},\text{i},\text{r},\text{e},\text{s},\text{sp},\text{pi},\text{ir},\text{re},\text{es},\text{spi},\text{pir},\text{ire},\text{res}\}$. This method of encoding variable length sequences is similar to \emph{Wickelphone} phoneme-encoding methods~\cite{Wickelgran69}.

Even for small values of $N$, $G_N(w)$ encodes each word $w\in\mathcal{W}$ nearly uniquely. For example, with $N=4$, this map has only 7 collisions out of a dictionary of 90k words. The encoding $G_N(w)$ can be represented as a $|G_N|$-dimensional binary vector of gram occurrences. This vector is very sparse, as on average $|G_N(w)|\approx 22$ whereas $|G_N|=10k$. Given $w$, we predict this vector using the same base CNN as in~\sref{sec:dictionary} and~\sref{sec:characters}, but now have a final fully connected layer with $|G_N|$ neurons to represent the encoding vector. The scores from the fully connected layer can be interpreted as probabilities of an N-gram being present in the image by applying the logistic function to each neuron. The CNN is therefore learning to recognise the presence of each N-gram somewhere within the input image.

\paragraph{Training.}
With a logistic function, the training problem becomes that of $|G_N|$ separate binary classification tasks, and so we back-propagate the logistic regression loss with respect to each N-gram class independently. To jointly train a whole range of N-grams, some of which occur very frequently and some barely at all, we have to scale the gradients for each N-gram class by the inverse frequency of their appearance in the training word corpus. We also experimented with hinge loss and simple regression to train but found frequency weighted binary logistic regression was superior. As with the other models, we use dropout and SGD.

% --------------------------------------------------------------
% --------------------------------------------------------------
\section{Evaluation}\label{sec:eval}
% --------------------------------------------------------------
% --------------------------------------------------------------
This section evaluates our three text recognition models.~\sref{sec:data} describes the benchmark data,~\sref{sec:implementation} the implementation details, and~\sref{sec:experiments} the results of our methods, that improve on the state of the art.
%Finally we show how the N-gram encoding model can be used for full word recognition, and the full text-spotting result for an example image.

\subsection{Datasets}\label{sec:data}
A number of standard datasets are used for the evaluation of our systems -- ICDAR 2003, ICDAR 2013, Street View Text, and IIIT5k. {\bf ICDAR 2003}~\cite{ICDAR03} is a scene text recognition dataset, with the test set containing 251 full scene images and 860 groundtruth cropped images of the words contained with the full images. We follow the standard evaluation protocol by~\cite{Wang11,Wang12,Alsharif13} and perform recognition on only the words containing only alphanumeric characters and at least three characters. The test set of 860 cropped word images is referred to as IC03. The lexicon of all test words is IC03-Full, and the per-image 50 word lexicons defined by~\cite{Wang11} and used in~\cite{Wang11,Wang12,Alsharif13} are referred to as IC03-50. There is also the lexicon of all groundtruth test words -- IC03-Full which contains 563 words. {\bf ICDAR~2013}~\cite{ICDAR2013} test dataset contains 1015 groundtruth cropped word images from scene text. Much of the data is inherited from the ICDAR 2003 datasets. We refer to the 1015 groundtruth cropped words as IC13. {\bf Street View Text}~\cite{Wang11} is a more challenging scene text dataset than the ICDAR datasets. It contains 250 full scene test images downloaded from Google Street View. The test set of 647 groundtruth cropped word images is referred to as SVT. The lexicon of all test words is SVT-Full (4282 words), and the smaller per-image 50 word lexicons defined by~\cite{Wang11} and used in~\cite{Wang11,Wang12,Alsharif13,Bissacco13} are referred to as SVT-50. {\bf IIIT 5k-word}~\cite{Mishra12} test dataset contains 3000 cropped word images of scene text downloaded from Google image search. Each image has an associated 50 word lexicon (IIIT5k-50) and 1k word lexicon (IIIT5k-1k).

For training, validation and large-lexicon testing we generate datasets using the synthetic text engine from~\sref{sec:synth}. 4 million word samples are generated for the IC03-Full and SVT-Full lexicons each, referred to as Synth-IC03 and Synth-SVT respectively. In addition, we use the dictionary from Hunspell, a popular open source spell checking system, combined with the ICDAR and SVT test words as a 50k word lexicon. The 50k Hunspell dictionary can also be expanded to include different word endings and combinations to give a 90k lexicon. We generate 9 million images for the 50k word lexicon and 9 million images for the 90k word lexicon. The 9 million image synthetic dataset covering 90k words, Synth, is available for download at {\small\url{http://www.robots.ox.ac.uk/~vgg/data/text/}}.

\subsection{Implementation Details}\label{sec:implementation}
We perform experiments on all three encoding models described in~\sref{sec:architectures}. We will refer to the three models as DICT, CHAR, and NGRAM for the dictionary encoding model, character sequence encoding model, and N-gram encoding model respectively. The input images to the CNNs are greyscale and resized to $32 \times 100$ without aspect ratio preservation. 
The only preprocessing, performed on each sample individually, is the sample mean subtraction and standard deviation normalization (after resizing), as this was found to slightly improve performance.
%
%We found experimentally that standard deviation normalization is not essential, but gives slightly better classification performance. No data jittering or random crops are needed due to the adequate size of the training dataset. Also at test time, no augmentation of test samples is performed, and there is no model averaging involved.
% It is likely that incorporating these tricks would increase performance slightly.
%
Learning uses a custom version of \texttt{Caffe}~\cite{Jia13}.

All CNN training is performed solely on the Synth training datasets, with model validation performed on a 10\% held out portion. The number of character classifiers in the CHAR character sequence encoding models is set to 23 (the length of the largest word in our 90k dictionary). In the NGRAM models, the number of N-grams in the N-gram classification dictionary is set to 10k. The N-grams themselves are selected as the N-grams with at least 10 appearances in the 90k word corpus -- this equates to 36 1-grams (the characters), 522 2-grams, 3965 3-grams, and 5477 4-grams, totalling 10k.

In addition to the CNN model defined in~\sref{sec:architectures}, we also define larger CNN, referred to as DICT+2, CHAR+2, and NGRAM+2. The larger CNN has an extra $3 \times 3$ convolutional layer with 512 filters before the final pooling layer, and an extra 4096 unit fully connected layer after the original 4096 unit fully connected layer. Both extra layers use rectified linear non-linearities. Therefore, the total structure for the DICT+2 model is conv-pool-conv-pool-conv-conv-pool-conv-fc-fc-fc, where conv is a convolutional layer, pool is a max-pooling layer and fc is a fully connected layer. We train these larger models to investigate the effect of additional model capacity, as the lack of over-fitting experienced on the basic models is suspected to indicate under-capacity of the models.

\subsection{Experiments}\label{sec:experiments}

%%%%%%%%%%%%%%%%%%%
%%% INTERNAL RESULTS
%%%%%%%%%%%%%%%%%%%
\begin{table}[t]
\begin{center}\scriptsize %\footnotesize
\setlength{\tabcolsep}{3pt}
\begin{tabular}[t]{|l|l||c|c|c|c|c|c|} 
%%%%%% Title row starts here
\hline
\multicolumn{1}{|c|}{\centering Model} & 
\multicolumn{1}{p{1.0cm}||}{\centering Trained Lexicon} &
\multicolumn{1}{p{1.2cm}|}{\centering Synth} &
\multicolumn{1}{c|}{\centering IC03-50} &
\multicolumn{1}{c|}{\centering IC03} &
\multicolumn{1}{c|}{\centering SVT-50} &
\multicolumn{1}{c|}{\centering SVT} &
\multicolumn{1}{c|}{\centering IC13}
\\
\hline\hline
%%%%%% Row ICDAR2003 starts here
DICT-IC03-Full & IC03-Full & 98.7 & 99.2 &  98.1 & - & - & - \\
\rowcolor{Gray}
DICT-SVT-Full & SVT-Full  & 98.7 & - & - & 96.1 & 87.0 & -\\
DICT-50k & 50k       & 93.6 & 99.1 & 92.1 & 93.5 & 78.5 & 92.0 \\ 
\rowcolor{Gray}
DICT-90k & 90k       & 90.3 & 98.4 & 90.0 & 93.7 & 73.0 & 86.3 \\ 
DICT+2-90k & 90k     & 95.2 & 98.7 & 93.1 & 95.4 & 80.7 & 90.8 \\
\rowcolor{Gray}
CHAR & 90k       & 71.0 & 94.2 & 77.0 & 87.8 & 56.4 & 68.8 \\ 
CHAR+2 & 90k     & 86.2 & 96.7 & 86.2 & 92.6 & 68.0 & 79.5 \\
\rowcolor{Gray}
NGRAM-NN & 90k      & 25.1 & 92.2 & - & 84.5 & - & -\\
NGRAM+2-NN & 90k    & 27.9 & 94.2 & - & 86.6 & - & - \\
%\rowcolor{Gray}
%CHARGRAM & 90k     & td & 96.5 & 89.5 & 92.6 & 68.0 & 79.5 & 95.5 & 85.4 & 72.3\\
\hline
\end{tabular}
\qquad\quad
\vtop{\vspace{0pt}\hbox{\includegraphics[width=0.23\textwidth]{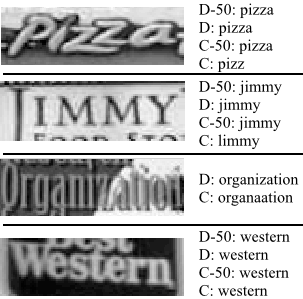}}}%
\end{center}
\vspace*{-1em}
\caption{\small \emph{Left:} The word recognition accuracy for the different proposed models with different trained lexicons. Where a lexicon is not specified for a dataset, the only language constraints are those imposed by the model itself. The fixed lexicon CHAR model results (IC03-50 and SVT-50) are obtained by selecting the lexicon word with the minimum edit distance to the predicted character sequence. 
\emph{Right:} Some random example results from the SVT and ICDAR 2013 dataset. D denotes DICT+2-90k with no lexicon, D-50 the DICT+2-90k model constrained to the image's 50 word lexicon, C denotes the CHAR+2 model with completely unconstrained recognition, and C-50 gives the result of the closest edit distance 50-lexicon word.} 
\label{table:internal}
\end{table}

We evaluate each of our three models on challenging text recognition benchmarks. First, we measure the accuracy on a large dataset, containing the images of words from the full lexicon (up to 90k words depending on the model). Due to the lack of human-annotated natural image datasets of such scale, we use the test split of our Synth dataset (Sect.~\ref{sec:data}). This allows us to assess how well our models can discriminate between a large number of words. Second, we consider the standard benchmarks IC03~\cite{ICDAR03}, SVT~\cite{Wang11}, and IC13~\cite{ICDAR2013}, which contain natural scene images, but cover smaller word lexicons. The evaluation on these datasets allows for a fair comparison against the state of the art. The results are shown in~\tblref{table:internal} and~\tblref{table:comparison}. 

\paragraph{Dictionary Encoding.}
For the DICT model, we train a model with only the words from the IC03-Full lexicon (DICT-IC03-Full), a model with only the words from the SVT-Full lexicon (DICT-SVT-Full), as well as models for the 50k and 90k lexicons -- DICT-50k, DICT-90k, and DICT+2-90k. When a small lexicon is provided, we set the language prior $P(w|\mathcal{L})$ to be equal probability for lexicon words, otherwise zero. In the absence of a small lexicon, $P(w|\mathcal{L})$ is simply the frequency of word $w$ in a corpus (we use the opensubtitles.org English corpus) normalized according to the power law.

The results in~\tblref{table:internal} show exceptional performance for the dictionary based models. When the model is trained purely for a dataset's corpus of words (DICT-IC03-Full and DICT-SVT-Full), the 50-lexicon recognition problem is largely solved for both ICDAR 2003 and SVT, achieving 99.2\% and 96.1\% word recognition accuracy respectively, that is 7 mistakes out of 860 in the ICDAR 2003 test set, of which most are completely illegible. 
The Synth dataset performs very closely to that of the ICDAR 2003 dataset, confirming that the synthetic data is close to the real world data.

Drastically increasing the size of the dictionary to 50k and 90k words gives very little degradation in 50-lexicon accuracy. However without the 50-lexicon constraint, as expected the 50k and 90k dictionary models perform significantly worse than when the dictionary is constrained to only the groundtruth words -- on SVT, the word classification from only the 4282 groundtruth word set yields 87\% accuracy, whereas increasing the dictionary to 50k reduces the accuracy to 78.5\%, and the accuracy is further reduced to 73.0\% with 90k word classes. Incorporating the extra layers in to the network with DICT+2-90k increases the accuracy a lot, giving 80.7\% on SVT for full 90k-way classification, almost identical to a dictionary of 50k with the basic CNN architecture.

\begin{figure}[t]
\begin{center}
\begin{tabular}{cc}
\includegraphics[width=0.45\textwidth]{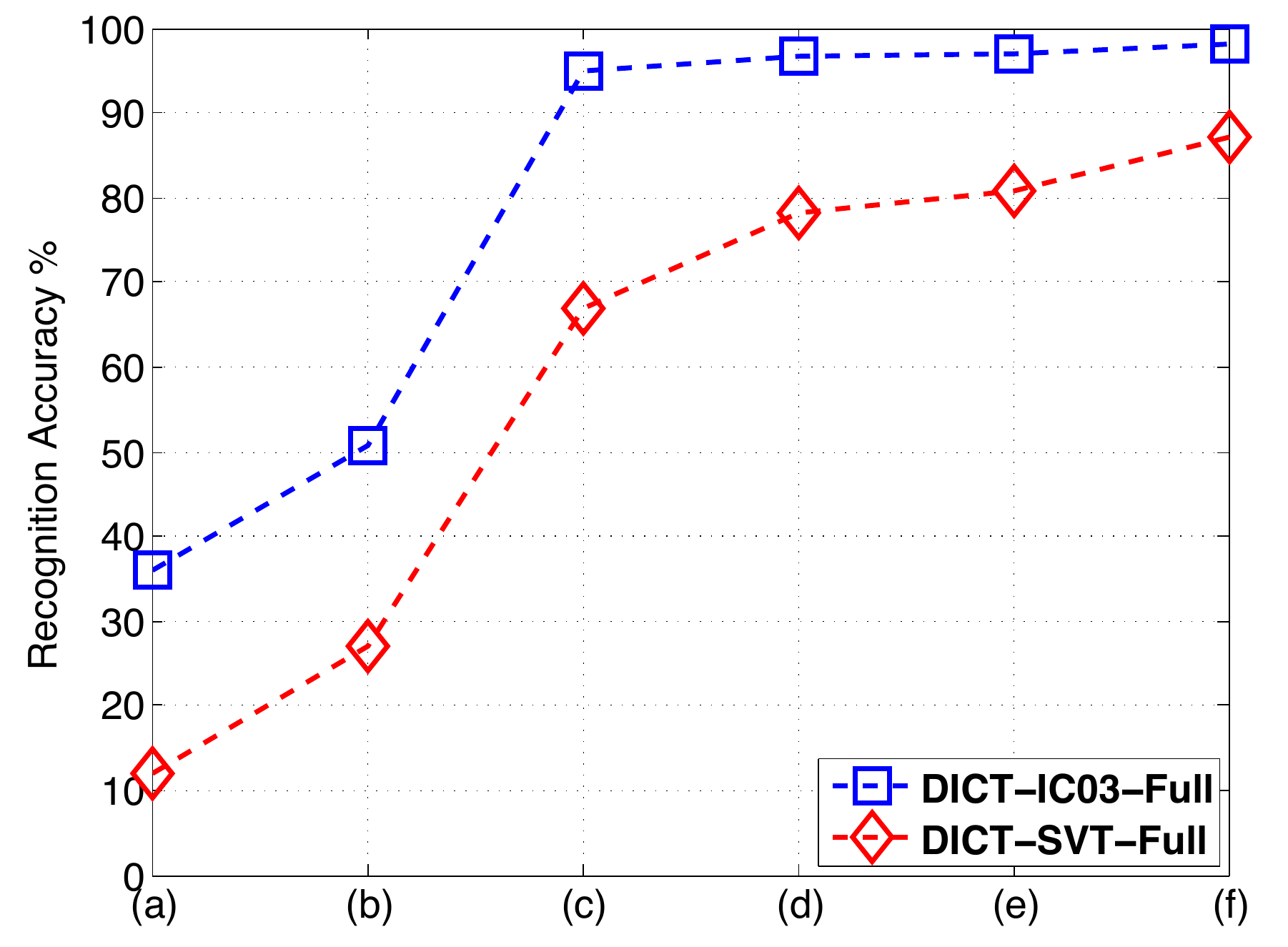}&
~~~~~~~\includegraphics[width=0.35\textwidth]{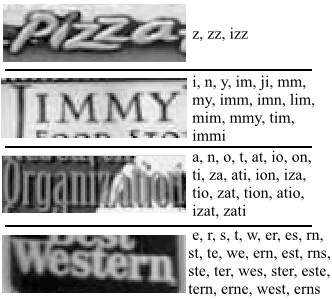}
\end{tabular}
\end{center}
\caption{\small \emph{Left:} The recognition accuracies of the DICT-IC03-Full and DICT-SVT-Full models evaluated on IC03 and SVT respectfully. The models (a-f) are trained on purely synthetic data with increasing levels of sophistication of the synthetic data. (a) Black text rendered on a white background with a single font, Droid Sans. (b) Incorporating all of Google fonts. (c) Adding background, foreground, and border colouring. (d) Adding perspective distortions. (e) Adding noise, blur and elastic distortions. (f) Adding natural image blending -- this gives an additional 6.2\% accuracy on SVT. \emph{Right:} The N-gram recognition results with probability over 0.99 from the NGRAM+2 model on random test images from SVT and ICDAR 2013.}
\vspace{0.5cm}
\label{fig:ngramresults}
\end{figure}

We also investigate the contribution that the various stages of the synthetic data generation engine make to real-world recognition accuracy. Figure~\ref{fig:ngramresults} (\emph{left}) shows DICT-IC03-Full and DICT-SVT-Full accuracy when trained identically but with different levels of sophistication of synthetic training data. As more sophisticated training data is used, the recognition accuracy increases -- the addition of random image-layer colouring causing a significant increase in performance (+44\% on IC03 and +40\% on SVT), as does the addition of natural image blending (+1\% on IC03 and +6\% on SVT).

%%%%%%%%%%%%%%%%%%%
%%% COMPARISON RESULTS
%%%%%%%%%%%%%%%%%%%
\setlength{\tabcolsep}{3pt}
\begin{table}[t]
\begin{center}
\scriptsize
\begin{tabular}[t]{|l||c|c|c|c|c|c|c|c|} 
%%%%%% Title row starts here
\hline
\multicolumn{1}{|c||}{\centering Model} & 
\multicolumn{1}{c|}{\centering IC03-50} &
\multicolumn{1}{c|}{\centering IC03-Full} &
\multicolumn{1}{c|}{\centering IC03-50k} &
\multicolumn{1}{c|}{\centering SVT-50} &
\multicolumn{1}{c|}{\centering SVT} &
\multicolumn{1}{c|}{\centering IC13} &
\multicolumn{1}{c|}{\centering IIIT5k-50} &
\multicolumn{1}{c|}{\centering IIIT5k-1k} \\
\hline\hline
%%%%%% Row ICDAR2003 starts here
\rowcolor{Gray}
\emph{Baseline ABBYY}~\cite{Wang11} & 56.0 & 55.0 & - & 35.0 & - & - & 24.3 & -\\
Wang~\cite{Wang11}          & 76.0 & 62.0 & - & 57.0 & - & - & - & -\\
\rowcolor{Gray}
Mishra~\cite{Mishra12}      & 81.8 & 67.8 & - & 73.2 & - & - & - & -\\
Novikova~\cite{Novikova12}  & 82.8 & - & - & 72.9 & - & - & 64.1 & 57.5\\
\rowcolor{Gray}
Wang~\&~Wu~\cite{Wang12}    & 90.0 & 84.0 & - & 70.0 & - & - & - & -\\
Goel~\cite{Goel13}          & 89.7 & - & - & 77.3 & - & - & - & -\\
\rowcolor{Gray}
PhotoOCR~\cite{Bissacco13}& - & - & - & 90.4 & 78.0 & 87.6 & - & -\\
Alsharif~\cite{Alsharif13}  & 93.1 & 88.6 & 85.1 & 74.3 & - & - & - & -\\
\rowcolor{Gray}
Almazan~\cite{Almazan14} & - & - & - & 89.2 & - & - & 91.2 & 82.1\\
Yao~\cite{Yao14}  & 88.5 & 80.3 & - & 75.9 & - & - & 80.2 & 69.3\\
\rowcolor{Gray}
Jaderberg~\cite{Jaderberg14a}  & 96.2 & 91.5 & - & 86.1 & - & - & - & -\\
Gordo~\cite{Gordo14} & - & - & - & 90.7 & - & - & 93.3 & 86.6\\
\hline
\rowcolor{Gray}
DICT-IC03-Full & \bf 99.2 & \bf 98.1 & - & - & - & - & - & -\\
DICT-SVT-Full  & - & - & - & \bf 96.1 & \bf 87.0 & - & - & -\\
%\rowcolor{Gray}
%DICT-50k       & \bf 99.1 & \bf 98.0 & \bf 92.1 & \bf 93.5 & \bf 78.5 & \bf 92.0\\
%DICT-90k       & \bf 98.4 & \bf 98.6 & TODO & \bf 93.7 & 73.0 & 86.3\\
\rowcolor{Gray}
DICT+2-90k     & \bf 98.7 & \bf 98.6 & \bf 93.3 & \bf 95.4 & \bf 80.7 & \bf 90.8 & \bf 97.1 & \bf 92.7\\
%CHAR       & \bf 94.2 & \bf 92.0 & TODO & 87.8 & 56.4 & 68.8\\
CHAR+2     & \bf 96.7 & \bf 94.0 & \bf 89.5 & \bf 92.6 & 68.0 & 79.5 & \bf 95.5 & 85.4\\
%NGRAM-90k      & TODO & TODO & TODO & TODO & TODO & TODO\\
%\rowcolor{Gray}
\rowcolor{Gray}
NGRAM+2-SVM & \bf 96.5 & \bf 94.0 & - & - & - & - & - & -\\
%CHARGRAM     & \bf 96.5 & \bf 95.0 & \bf td & \bf 91.8 & 70.5 & 81.5 & - & -\\

\hline
\end{tabular}
\end{center}
\vspace*{-1em}
\caption{\small Comparison to previous methods. The ICDAR 2013 results given are case-insensitive. Bolded results outperform previous state-of-the-art methods. The baseline method is from a commercially available OCR system. }
\label{table:comparison}
\end{table}

\paragraph{Character Sequence Encoding.}
The CHAR models are trained for character sequence encoding. The models are trained on image samples of words uniformly sampled from the 90k dictionary.

The output of the model are character predictions for a possible 23 characters of the test image's word. We take the predicted word as the MAP-optimal sequence of characters, stripping any no-character classifications. The constrained lexicon results for IC03-50, IC03-Full, and SVT-50, are obtained by finding the lexicon word with the minimum edit distance to a raw predicted character sequence. Given this is a completely unconstrained recognition, with no language model at all, the results are surprisingly good. The 50-lexicon results are very competitive compared to the other encoding methods. However, we can see the lack of language constraints cause the out-of-lexicon results to be lacklustre, achieving an accuracy of only 79.5\% with the CHAR+2 model on ICDAR 2013 as opposed to 90.8\% with the DICT+2-90k model. As with the DICT models, increasing the number of layers in the network increases the word recognition accuracy by between 6-8\%.

Some example word recognition results with dictionary and character sequence encodings are shown to the right of~\tblref{table:internal}.

\paragraph{Bag-of-N-grams Encoding.}
The NGRAM model's output is thresholded to result in a binary activation vector of the presence of any of 10k N-grams in a test word.
%an activation vector of the presence of each of the 10k dictionary N-grams found in the test sample. Due to the variable number of N-grams found in a word, the sparse activation vector must be thresholded to infer an explicit list of predicted N-grams. 
%
%Shown in~\figref{fig:ngrampr} is the PR curve for N-gram recognition, with a maximum harmonic mean of {\bf XXX} achieved.
%
Decoding the N-gram activations into a word could take advantage of a statistical model of the language. Instead, 
%In the presence of a language the N-grams can easily be used to infer a dictionary word, but this is beyond the scope of this paper. However, 
we simply search for the word in the lexicon with the nearest (in terms of the Euclidean distance) N-gram encoding, denoted as NGRAM-NN and NGRAM+2-NN models. This extremely naive method still gives competitive performance, illustrating the discriminative nature of N-grams for word recognition. Instead, one could learn a linear SVM mapping from N-gram encoding to dictionary words, allowing for scalable word recognition through an inverted index of these mappings. We experimented briefly with this on the IC03-Full lexicon -- training an SVM for each lexicon word from a training set of Synth data, denoted as NGRAM+2-SVM -- and achieve 97\% accuracy on IC03-50 and 94\% accuracy on IC03-Full. Figure~\ref{fig:ngramresults} (\emph{right}) shows the N-gram recognition results for the NGRAM+2 model, thresholded at 0.99 probability.

\paragraph{Comparison \& Discussion.}
\tblref{table:comparison} compares our models to previous work, showing that all three models achieve state-of-the-art results in different lexicon scenarios. 
With tightly constrained language models such as in DICT-IC03-Full and DICT-SVT-Full, we improve accuracy by +6\%. 
However, even when the models are expanded to be mostly unconstrained, such as with DICT+2-90k, CHAR+2 and NGRAM+2-SVM, our models still outperform previous methods. Considering a complete absence of a language model, the no-lexicon recognition results for the CHAR+2 model on SVT and IC13 are competitive with the system of~\cite{Bissacco13}, and as soon as a language model is introduced in the form of a lexicon for SVT-50, the simple CHAR+2 model gives +2.2\% accuracy over~\cite{Bissacco13}.
Performance could be further improved by techniques such as model averaging and test-sample augmentation, albeit at a significantly increased computational cost. Our largest model, the DICT+2-90k model comprised of over 490~million parameters, can process a word in 2.2ms on a single commodity GPU.

Our models set a new benchmark for scene text recognition. In a real-world system, the large DICT+2-90k model should be used for the majority of recognition scenarios unless completely unconstrained recognition is required where the CHAR+2 model can be used. However, when looking at the average edit distance of erroneous recognitions, the CHAR+2 model greatly outperforms the DICT+2-90k model, with an average error edit distance of 1.9 compared to 2.5 on IC13, suggesting the CHAR+2 model may be more suitable for a retrieval style application in conjunction with a fuzzy search.

\section{Conclusion}\label{sec:conclusions}
In this paper we introduced a new framework for scalable, state-of-the-art word recognition -- synthetic data generation followed by whole word input CNNs. We considered three models within this framework, each with a different method for recognising text, and demonstrated the vastly superior performance of these systems on standard datasets. In addition, we introduced a new synthetic word dataset, orders of magnitude larger than any released before. 
% There are many extensions that are possible to these models taking further advantage of this plethora of data. For future work we hope to combine the three models of different reading styles together in a multi-task scenario, creating a unified recognition system.

\section*{Acknowledgements}
This work was supported by the EPSRC and ERC grant VisRec no. 228180. We gratefully acknowledge the support
of NVIDIA Corporation with the donation of the GPUs used for this research.

\bibliographystyle{plainnat}
{
\small
\bibliography{bib/shortstrings,bib/vgg_local,bib/vgg_other,bib/current}
}
\end{document}